# Computer Assisted Composition with Recurrent Neural Networks

**Christian Walder**  christian.walder@data61.csiro.au
*Data61 at CSIRO, Australia.*

**Dongwoo Kim**  dongwoo.kim@anu.edu.au
*Australian National University.*

## Abstract

Sequence modeling with neural networks has lead to powerful models of symbolic music data. We address the problem of exploiting these models to reach creative musical goals, by combining with human input. To this end we generalise previous work, which sampled Markovian sequence models under the constraint that the sequence belong to the language of a given finite state machine provided by the human. We consider more expressive non-Markov models, thereby requiring approximate sampling which we provide in the form of an efficient sequential Monte Carlo method. In addition we provide and compare with a beam search strategy for conditional probability maximisation.

Our algorithms are capable of convincingly re-harmonising famous musical works. To demonstrate this we provide visualisations, quantitative experiments, a human listening test and audio examples. We find both the sampling and optimisation procedures to be effective, yet complementary in character. For the case of highly permissive constraint sets, we find that sampling is to be preferred due to the overly regular nature of the optimisation based results. The generality of our algorithms permits countless other creative applications.

**Keywords:** Music, sequence models, composition, neural networks.

## 1. Introduction

Algorithmic music composition has intrigued a wide range of thinkers, from times as distant as the earliest days of modern computing, and beyond. As early as 1843, Ada Lovelace famously speculated that a computer "might compose elaborate and scientific pieces of music of any degree of complexity or extent" (Menabrea and Lovelace, 1843). Still earlier, around 1757, there famously existed a number of musical games which used dice along with precise instructions, to generate musical compositions (Nierhaus, 2008).

Much work has focused on machine learning methods, which combine musical examples with generic inductive principles, to generate new examples. For a survey of more recent advancements we recommend *e.g.* Fernandez and Vico (2013); Nierhaus (2008). By restricting to a highly structured and regular musical form, and providing a large number of training examples, we may obtain convincing musical results even with rather music-agnostic models (Sturm et al., 2015). Alternatively, with smaller datasets, more structured models are required, such as the hidden Markov model investigated by Allan and Williams (2005). Both of these methods lead to relatively convincing musical results. However, the plausibility of the results comes at the cost of variety. It is unreasonable to expect such approaches to give rise to interesting new musical forms.

Unmanned machine learning algorithms capable of divining interesting new musical forms may arise, but this is far from a reality at the time of writing. In any case, hybrid man-machine systems are a promising avenue for exploration. The high level goal is a system for the partial specification of music, which relinquishes precise control for other gains.[1]

In this work, we investigate the combination of machine learning music models with human input. Our approach is most closely related to the work of Papadopoulos et al. (2015), who impose the constraint that the resulting sequence be a member of a language defined by a finite state machine. This is a rather general constraint which provides a rich language for human creativity via the partial specifications of music. In that work, the close relationship between Markov models and finite state machines was used to derive a simple but exact belief propagation algorithm. Here, we relax the constraint that our underlying sequence model be Markovian, and instead make use of approximate sampling techniques, namely that of sequential Monte Carlo (SMC) (Gordon et al., 1993; Doucet and Johansen, 2009). In addition to sampling, we compare and contrast with a maximum conditional probability approach based on a beam search. (given the finite state machine constraint).

By allowing models which are not low-order Markovian, we obtain algorithms which are applicable to a broad range of more expressive probabilistic models of music. Concretely, here we adopt the Long Short-Term Memory (LSTM) based neural network sequence model (Hochreiter and Schmidhuber, 1997). By exploiting advances in massively parallel computing architectures, this class of models has proven remarkably effective in a range of domains, most notably natural language modeling (see *e.g.* Chelba et al. (2013)). Given the similarities between music and natural language (Patel, 2010), it is unsurprising that the LSTM is a natural model for music data. This was noted by Hochreiter and Schmidhuber (1997) in the original LSTM paper, and subsequently investigated by Eck and Schmidhuber (2002). This line of work has been taken up by a number of researchers, with recent examples including *e.g.* Boulanger-Lewandowski et al. (2012); Walder (2016), the latter approach being the one we adopt here.

In section 2 we set the problem up and motivate our solution. We provide technical details of the approximate sampling method in section 3. Section 4 provides results including a visual illustration, quantitative investigation, human listening test, and audio examples. We conclude in section 5.

## 2. Set-up and Motivation

### 2.1. Required Background

We take the model of Walder (2016) as our starting point, and while this work is a necessary accompaniment to the present work for those wishing to replicate our results, the present paper stands alone in the sense that it may be applied to any model which factorises in the causal manner subsection 2.2 described below. Furthermore, the finite state machine constraint formalism we adopt is closely related to Papadopoulos et al. (2015), which we also recommend as background material.

---

1. *I will argue that today's composers are more frequently gardeners than architects and, further, that the composer as architect metaphor was a transitory historical blip* — Eno (2011).

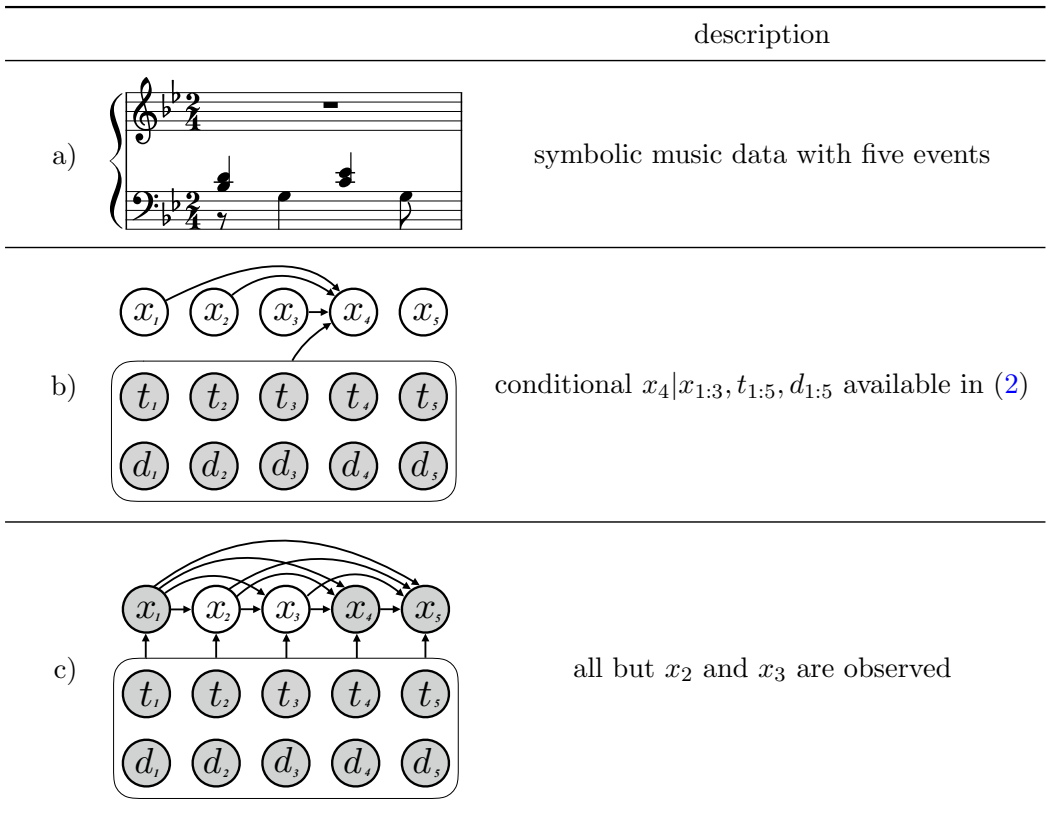

Figure 1: An overview of our system: please refer to section 2 for more details on the background. Symbolic music consisting of five events (row a — note that the *rests* or silences are not counted) is represented by a sequence of five note names (roughly speaking, indices into a piano keyboard) $x_{1:5}$, along with start times $t_{1:5}$ and durations $d_{1:5}$. Following Walder (2016) we model the conditional $x_{1:5}|t_{1:5}, d_{1:5}$, with a causal factorisation over $x_{1:5}$ as indicated by the diagram (row b), and the equation (2). Row c depicts the simplest setting we consider, wherein the human input amounts to the partial observation of the sequence $x_{1:5}$ (see subsection 3.1). In the case depicted in row c, we infer some fixed subset of the notes (here, $x_2$ and $x_3$), given the values for all the other variables. The resulting conditional $x_2, x_3|x_1, x_4, x_5, t_{1:5}, d_{1:5}$ is non-trivial due to observed future values, which are not directly handled by the conditionals which we explicitly model via the causal factorisation (2). In rows b–c the box around $t_{1:5}$ and $d_{1:5}$ is meant to denote that the out-going arrow(s) emanate from all the variables in the box.

## 2.2. Assumed Music Model

We represent a musical composition by a set of $n$ triples, $\{(x_i, t_i, d_i)\}_{i=1}^n \subset \mathcal{X} \times \mathcal{T} \times \mathcal{T}$. Here, $\mathcal{X}$ represents the set of possible pitches (from the 12-tet western musical system, including both the note name and octave). The set $\mathcal{T}$ represents time, with $t_i$ and $d_i$ denoting the start time and duration of the $i$-th note event, respectively.

Following Walder (2016), we assume throughout that the timing information is given, so that we have a model $p(\{x_i\}_{i=1}^n \mid n, \{(t_i, d_i)\}_{i=1}^n)$. This model $p$ is based on a reduction to a non-Markov sequence model. This means that (for some particular ordering of the indices, see Walder (2016)) neglecting the conditioning on $\{(t_i, d_i)\}_{i=1}^m$, we have

$$p(x_{1:n}) = \prod_{i=1}^n f_i(x_i | x_{1:i-1}), \qquad (1)$$

where the conditionals $f_i$ are represented explicitly.

## 2.3. Assumption of Fixed Rhythmic Information

Ideally we would model the rhythmic structure, but this turns out to be rather challenging. Indeed, modeling pitches given the rhythmic structure is already rather non-trivial, so it is reasonable to subdivide the problem. Note that some authors have modeled timing information, by assuming *e.g.*

1. A simple note on/off representation on a uniform temporal grid as in *e.g.* Boulanger-Lewandowski et al. (2012), which assumed a temporal resolution of an eighth note. This has the drawback of not distinguishing, say, two eighth notes from a single quarter note (which has length two eighth notes).

2. Rather than explicitly processing each time step on a fixed temporal grid, one may model time increments as in Colombo et al. (2016); Simon and Oore (2017). In this case sampling conditionally given constraints as is our present focus becomes more technical. Indeed, point process machinery (Daley and Vere-Jones, 2003) is required, due to the variable dimension of the parameter space (that is, the number of notes). The sampling problems analogous to the present work become highly non-trivial in that case. This is the subject of ongoing work, and beyond the present scope.

Short of developing the second approach, assuming fixed timing has the advantage of allowing arbitrarily complex rhythmic structures. In this way, rather than fully modeling an overly simplistic music representation, we partially model a more realistic music representation. Once we have satisfactorily modeled timing, our methods and findings may be generalised to the full joint distribution, as we can always employ the chain rule of probability to decompose

$$p\left(n, \{x_i, t_t, d_i\}_{i=1}^n\right) = p(\{x_i\}_{i=1}^n \mid n, \{(t_i, d_i)\}_{i=1}^n) \times p(n, \{(t_i, d_i)\}_{i=1}^n), \qquad (2)$$

where the first factor on the r.h.s. is our present concern.

## 2.4. Finite State Machine Constraint Formulation

## 3. Sequential Monte Carlo Details

We describe detailed sequential sampling methods. In subsection 3.1 we consider partially observed sequences, and then in subsection 3.2 we extend this to more general constraints represented by an arbitrary finite state machine. Throughout this section we neglect conditioning on the start times $t_i$ and durations $d_i$, as they are fixed throughout.

### 3.1. Partially Observed Sequences

Let $x_{1:n}$ be a sequence drawn from the non-Markov process with discrete state-space $\mathcal{X}$.

$$p(x_{1:n}) \propto \gamma_n(x_{1:n}) = \prod_{i=1}^{n} f_i(x_i|x_{1:i-1}),$$

where $\gamma_n$ is an un-normalised probability distribution which can be factorised into the conditionals $f_i$, each of which depends on all of the previous states $x_{1:i-1}$.

Our goal is to compute the posterior of $x_{1:n}$ with partial observations $\tilde{x}$. Let $z_i$ be 1 if $x_i$ is observed and 0 otherwise. The posterior distribution of the unobserved part is:

$$p(\{x_i\}_{i:z_i=0}|\{\tilde{x}_j\}_{j:z_j=1}, z_{1:n}) = \frac{p(x_{1:n})}{p(\{x_j\}_{j:z_j=1})}$$
$$\propto \frac{\prod_{i=1}^{n} f_i(x_i|x_{1:i-1})}{\sum_{x_i \in \mathcal{X},\, \forall i:z_i=0} \prod_{i=1}^{n} f_i(x_i|x_{1:i-1})},$$

where we let $x_{1:n}$ be the full sequence with the partial observations. Even though our state space $\mathcal{X}$ is finite, naïvely evaluating the probability of a partially observed sequence would involve computing $|\mathcal{X}|^{\sum_i (1-z_i)}$ terms, so that the computational cost is exponential with respect to the number of unobserved notes.

To resolve this computational problem, we develop employ a standard particle filter approach (Doucet and Johansen, 2009, section 3.2). One can approximate the posterior of $x_{1:n}$ by importance sampling with fully factorised proposal distribution $q(x_{1:n}) = \prod_{i=1}^{n} q_i(x_i|x_{<i})$, so that

$$\gamma_n(x_{1:n}) = w_n(x_{1:n})q(x_{1:n})$$
$$p(x_{1:n}) = \frac{w_n(x_{1:n})q(x_{1:n})}{\sum_{x_i \in \mathcal{X},\, \forall i=1,2,\ldots,n} w_n(x_{1:n})q(x_{1:n})}.$$

With $S$ samples $x_{1:n}^{(s)}$ from $q(x_{1:n})$, the posterior can be approximated by an empirical measure of the samples:

$$p(x_{1:n}|z_{1:n}) = \sum_{s=1}^{S} \frac{w_n(x_{1:n}^{(s)})}{\sum_{s'=1}^{S} w_n(x_{1:n}^{(s')})} \delta_{x_{1:n}^{(s)}}(x_{1:n}).$$

The weight of each sample $x_{1:n}^{(s)} \sim q(x_{1:n})$ is:

$$
\begin{aligned}
w_n(x_{1:n}^{(s)}) &= \frac{\gamma_n(x_{1:n}^{(s)})}{q(x_{1:n}^{(s)})} \\
&= \frac{\prod_{i=1}^n f_i(x_i^{(s)}|x_{1:i-1}^{(s)})}{q(x_{1:n}^{(s)})} \\
&= \underbrace{\frac{\tilde{\gamma}_{n-1}(x_{1:n-1}^{(s)})}{q(x_{1:n-1}^{(s)})}}_{\triangleq w_{n-1}(x_{1:n-1}^{(s)})} \times \underbrace{\frac{f_n(x_n^{(s)}|x_{1:n-1}^{(s)})}{q_n(x_n^{(s)}|x_{<n}^{(s)})}}_{\triangleq w_n(x_n^{(s)})} .
\end{aligned}
\quad (3)
$$

The above recursion defines the particle filter scheme (Doucet and Johansen, 2009). For our particular case of partial observations $\tilde{x}$, we define the proposal

$$
q_i(x_i|x_{<i}, z_{1:i}) = \left[\delta_{\tilde{x}_i}(x_i)\right]^{z_i} \left[f_i(x_i|x_{1:i-1}^{(s)})\right]^{1-z_i}, \quad (4)
$$

so that our proposal always proposes $\tilde{x}_j$ whenever the sampler encounters the partial observation. Then from (3) we have that the weight of sample $x_i^{(s)} \sim q_i(x_i)$ is

$$
w_i(x_i^{(s)}) \propto \begin{cases} 1 & z_i = 0 \\ f_i(x_i^{(s)}|x_{1:i-1}^{(s)}) & z_i = 1 \end{cases} \quad (5)
$$

This expression is intuitive: we sample in temporal order using our original (unconstrained) model, and re-weight our particles only when encountering an observed symbol (or constraint), giving more weight to those particles whose history better agrees with the observed symbol, as measured by the conditional $f_i$. This algorithm is a special case of that of the following subsection, for which pseudo-code is provided as algorithm 1.

### 3.2. General Constraints

Here we extend the set-up of the previous section to the more general case of sampling with the hard constraint that the sampled sequence lie within a given regular language. In particular, we wish to sample from

$$
\begin{aligned}
p^{(\mathcal{A})}(x_{1:n}) &\triangleq p(x_{1:n}| x_{1:n} \in \mathcal{L}(\mathcal{A})) \\
&\propto \begin{cases} p(x_{1:n}) & x_{1:n} \in \mathcal{L}_n(\mathcal{A}), \\ 0 & \text{otherwise.} \end{cases}
\end{aligned}
\quad (6)
$$

Here, $\mathcal{L}_n(\mathcal{A})$ is the set of sequences of length $n$ generated by finite state machine $\mathcal{A} = \langle Q, \Sigma, \delta, q_0, F \rangle$, where $Q$ is a set of states, $\Sigma$ an alphabet of actions, $\delta$ the transition function linking a state $q \in Q$ and a label $a \in \Sigma$ to the successor state $q' = \delta(q, a)$, $q_0 \in Q$ the initial state, and $F \subseteq Q$ the set of terminal states. We have $\mathcal{X} = \Sigma$ in our case. Similarly to Papadopoulos et al. (2015) we rewrite

$$
p^{(\mathcal{A})}(x_{1:n}) \propto \prod_{i=1}^n f_i(x_i|x_{1:i-1}) h_i(x_{1:i}),
$$

where

$$h_i(x_{1:i}) = \begin{cases} 1 & x_{1:i} \in \mathcal{L}_i(\mathcal{A}), \\ 0 & \text{otherwise,} \end{cases}$$

Papadopoulos et al. (2015) transformed their similar problem into one involving the Cartesian product of the domain of the $x_i$ and the state space of $\mathcal{A}$ to obtain a tree structured factor graph for which belief propagation is efficient. We are interested in non-Markov $p$ and hence have nothing to gain from such a transformation. Instead, we simply encode the language membership constraint via the general functions $h_i$ and proceed with SMC as usual.[2]

Now, we generalise (4) of by choosing

$$q^{(\mathcal{A})}(x_i|x_{1:i-1}) = \frac{1}{Z_q(x_{1:i-1})} f_i(x_i|x_{1:i-1}) h_i(x_{1:i}), \tag{7}$$

where $Z_q(x_{1:i-1}) = \sum_{x_i \in \mathcal{X}} f_i(x_i|x_{1:i-1}) h_i(x_{1:i})$. The re-sampling weights are derived similarly to (3), yielding (up to an irrelevant constant to which SystematicResample (Carpenter et al., 1999) of algorithm 1 is invariant):

$$\begin{aligned} w_i^{(\mathcal{A})}(x_i^{(s)}) &= Z_q(x_{1:i-1}) \\ &= \sum_{\tilde{x}_i \in \mathcal{X}} f_i(\tilde{x}_i|x_{1:i-1}^{(s)}) h_i(\tilde{x}_{1:i}^{(s)}), \end{aligned} \tag{8}$$

where $\tilde{x}_{1:i}^{(s)}$ is the concatenation of $x_{1:i-1}^{(s)}$ and $\tilde{x}_i^{(s)}$. Hence (8) reduces to (5) for unary constraints. The above expression is intuitive — if all the $h_i(\tilde{x}_{1:i}^{(s)})$ are one, the weights are one so we sample from $p$ without modification. In other cases, we give greater weight to those particles for which the total probability mass of admissible continuations is greater. The overall procedure is summarised by algorithm 1.

For concreteness, we provide two examples of finite state machine formulation. It is not crucial to understand these examples in detail however, as we may also formulate our user constraints by specifying the $h_i$ of Equation 3.2 directly.

**Example 1 (Sequence with unary constraints)** *We impose unary constraints at each time $i$ of length-$n$ sequence. Take $\mathcal{A} = \langle Q, \Sigma, \delta, q_0, F \rangle$, where each time $i$ has one-to-one corresponding state $q_i$. Let $\mathcal{X}_i$ be a set of acceptable notes at time $i$. With transition function $\delta(q_i, x_{i+1}) = q_{i+1}$ if $x_{i+1} \in \mathcal{X}_{i+1}$, we only allow notes in $\mathcal{X}_{n+1}$ and rejects notes not in $\mathcal{X}_{i+1}$. Furthermore choose the terminal states $F$ to be $\{q_n\}$. Hence, machine $\mathcal{A}$ only accepts length-$n$ sequences which satisfy predefined unary constraints $\mathcal{X}_i$. For example, in the partially observed sequence case, $\mathcal{X}_i = \{\tilde{x}_i\}$ if $z_i = 1$ and $\mathcal{X}_i = \mathcal{X}$ if $z_i = 0$. Note that this partially observed constraint is a special case of general unary constraints where $\mathcal{X}_i$ consists of single observation $\tilde{x}_i$.*

**Example 2 (Prohibit repeated notes)** *One can disallow consecutive positions from having the same note as follows. Let the number of states be equal to the number of possible notes $|\mathcal{X}|$, and each state $q_x$ be indexed by $x$. Choose the transition function $\delta(q_{x_i}, x_j) = q_{x_j}$ if $x_i \neq x_j$. $\mathcal{A}$ only accepts sequences which does not have the same two consecutive notes.*

---

2. However, since we only ever append an $x_i$ to a given particle, we may leverage the finite state machine formulation of $\mathcal{A}$ in order to evaluate the $h_i$, typically in $O(1)$.

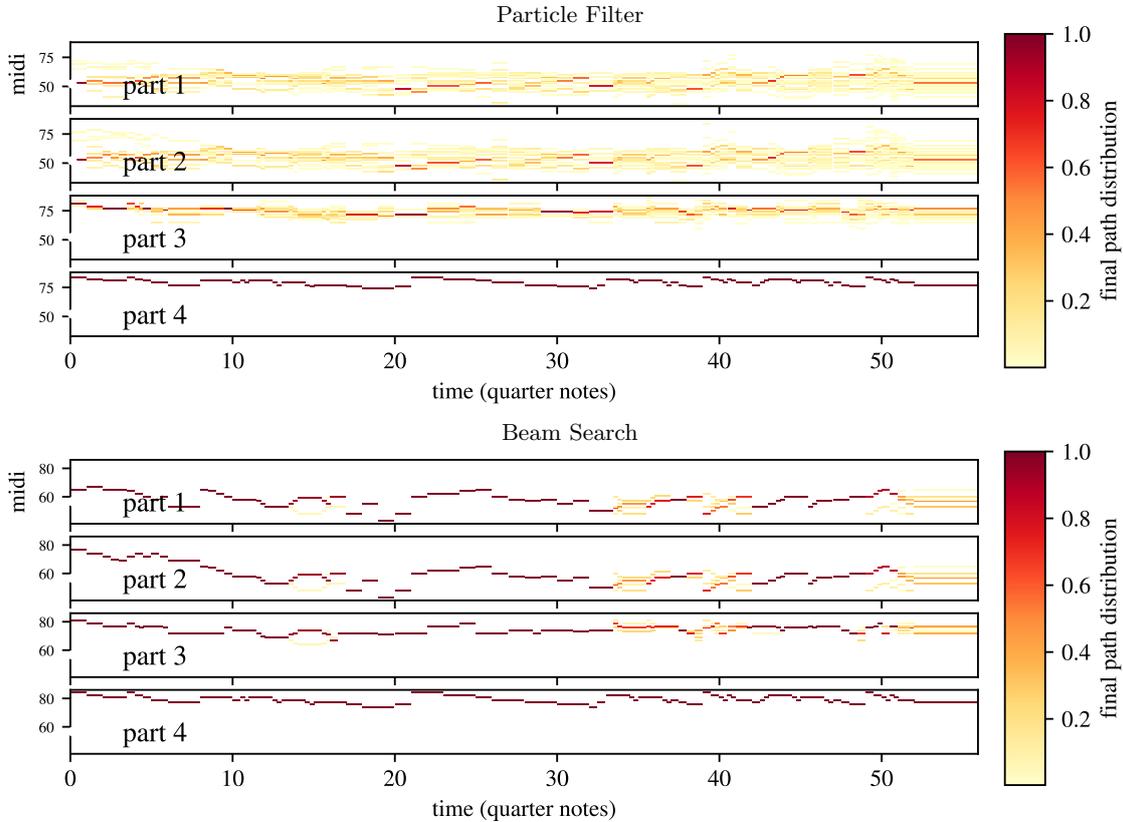

Figure 2: Visualisation of the final state of the particle filter (upper) and beam search (lower), with $S = 2048$ paths. In both cases the fourth part (lowest sub figure) is fixed to the original from Corelli's Trio Op 1 No 1 (which actually contains *four* midi tracks) provided by Boulanger-Lewandowski et al. (2012); hence the fourth part appears above as a deterministic distribution, with probability mass one on a single note at each time step. The pitches (vertical axis) of Parts 1–3 were unconstrained, to be determined by our algorithms. For the particle filter, the plotted distribution is a sequential Monte Carlo approximation of the filtering distribution (which, given that we consider the final time step, is also the smoothing distribution). For the beam search, the plotted distribution is simply the set of the candidate paths maintained by the deterministic search algorithm at its final step. Note that the part numbers, which corresponding to the axes above, and which come from the original composition, are used only for plotting and not by our algorithms.

Following Papadopoulos et al. (2015) we formulate human input as the constraint that the sequence $x_{1:n}$ belong to the language of an arbitrary finite state machine, $\mathcal{A}$. This has the advantage of being extremely general, yet amenable to belief propagation (for Markov $p$ as in Papadopoulos et al. (2015)), and particle filtering/beam search (as in the present work).

The experiments in section 4 utilise a fraction of the generality of this approach. We envisage a rich set of constraints which users may experiment with to discover interesting new music, and our algorithms facilitate this. One might enforce the repetition of patterns, transform relationships between sub-sequences (such as inversion / retrograde), and so on, with arbitrarily complex implications such as odd poly-metrical patterns, *etc.* Concrete examples are given by Roy and Pachet (2013), which imposes metrical structure, and Papadopoulos et al. (2016), which avoids the exact repetition of training sub-sequences.

### 3.3. Harmonisation by Sampling / Optimisation

We consider two alternative means of incorporating the constraint of the previous subsection. The first is conditional sampling, from $p(x_{1:n} | x_{1:n} \in \mathcal{L}(\mathcal{A}))$, where $\mathcal{L}(\mathcal{A})$ is the set of sequences $\mathcal{A}$ can generate. The second is maximisation of the same conditional probability with respect to $x_{1:n}$.

The nature of (1) is such that given a partial sequence $x_{1:i-1}$ we may keep track of a latent variable (the LSTM state vector) in order to compute $f_i(x_i|x_{1:i-1})$ in time independent of $i$. The state vector may also be updated in time independent of $i$, when a new element $x_i$ is appended. This structure makes SMC the natural choice for sampling, since it only ever requires extending a given sequence by a single element. Section 3 provides the details. This approach is in contrast to *e.g.* Hadjeres and Pachet (2017), wherein a simpler conditional independence structure is assumed (a local dependency network), making sampling simpler but precluding the modelling of longer range dependencies as captured by the LSTM.

Similarly, beam search methods are natural for maximising the conditional probability. In this case we maintain $S$ candidate paths, and extend each path by one new element each time step, for all feasible continuations, retaining the $S$ most probable resulting sequences.

### 4. Results

Throughout the experiments, we employ finite state machines which merely fix certain notes $x_i$ to given values, and prevent unison intervals from occurring within a single part (which would be impossible to play on the piano, for example).

### 4.1. Visualisation

A well known difficulty in particle filtering is the collapsing of the particles to a small number of unique paths. We observe little evidence of this in our setting however, which features a reasonable degree of uncertainty in the final particle distribution (Figure 2, upper plot). As expected, the beam search (Figure 2, lower plot) behaves very differently, with the majority of the solutions being identical other than at a few isolated temporal intervals.

### 4.2. Quantitative Investigation

Both the beam search and particle filter algorithms have a single key free parameter $S$, the number of paths to store, which trades accuracy for computation time. In this subsection, we investigate this trade-off. For computational reasons, we restricted our analysis to the sixteen shortest pieces (by number of notes) from those pieces in the *MuseData* test set from Boulanger-Lewandowski et al. (2012) which consist of exactly four voices (all of which turned out to be Bach chorales). For each piece, and for $m \in \{1, 2, 3, 4\}$, we fixed $m$ of the voices to that of the original piece, and applied our algorithms to choose the remaining $(4 - m)$ parts. We did this in all $C_m^4$ possible ways, plotting in Figure 3 the mean and standard error over all $16 \times C_m^4$ possibilities of two different quantities, as a function of the number of paths $S$. The two quantities depicted are:

1. Figure 3 l.h.s.: the filtering probability of the fixed parts — in the notation of subsection 3.2 the mean of the log of

$$\sum_{s=1}^{S} \frac{w(x_{<i}^{(s)})}{\sum_{s'=1}^{S} w(x_{<i}^{(s')})} f_i(x_i | x_{<i}^{(s)}),$$

for all $i$ corresponding to the fixed voices. For the particle filter, this is an SMC approximation of the probability of the ground truth for the fixed voices conditional on previous observed values for the fixed voices. For the beam search we let $w(x_{<i}^{(s)}) = 1$, and the quantity heuristically measures how well the set of candidate paths agree with the fixed voices on average.

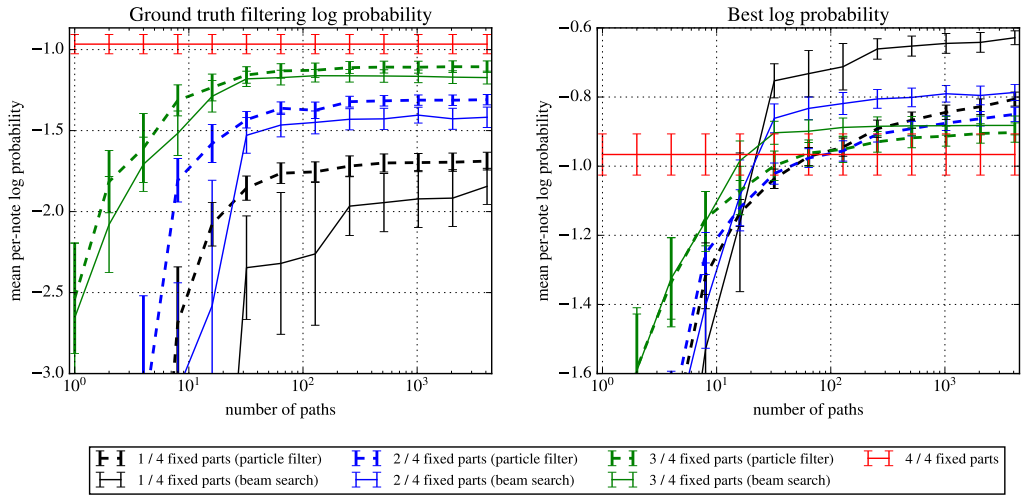

Figure 3: The effect of increasing the number of paths. Left: the filtering probability of the fixed (ground truth) parts, and right: the best (maximum) probability sequence found by the algorithm. The vertical axis is the mean log probability per musical note under our sequence model. The mean +/- one standard error of this quantity is depicted, taken over the 16 pieces we consider, multiplied by the $C_m^4$ choices of fixed parts, where the number $m$ of fixed parts ranges from 1 to 4 and is denoted by the color. The beam search is plotted in solid lines and the particle filter in broken lines. See subsection 4.2 for the details.

2. Figure 3 r.h.s.: the probability of the best complete harmonisation found by the algorithm. For the beam search, this is the log probability of the final path obtained

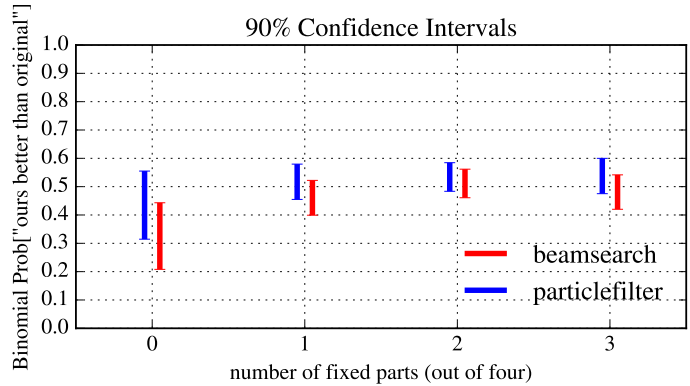

Figure 4: Results from the listening test described in subsection 4.3. We plot 90% confidence intervals for the probability that our results are preferred to the original compositions from which they were derived. The horizontal axis is the number of parts which were fixed to the values in the original composition. We see that with the provision of just one part from the original piece, we score approximately as well as the original composer (*i.e.* $p = 0.5$), namely Bach.

    by the search algorithm. For the particle filter, we take the most likely solution from the final set of particles.

Qualitatively, the results are as expected, with each algorithm dominating in the setting it is intended for: the particle filter is superior in terms of the filtering probability, while the beam search produces more probable solutions.

    Interestingly, given sufficiently large $S$, the best log probability exceeds the ground truth (4/4 fixed parts, red line). This indicates that either we are doing better than the original composer, or the probability under $p$ does not fully reflect perceived quality. The listening test in subsection 4.3 indicates that the latter is of course most likely, although this does not necessarily imply that $p$ is inadequate — see subsubsection 4.3.2 and footnote 4.

    Quantitatively, we see that around $S = 500$ paths are necessary in order to obtain good solutions (that the particle filter continues to improve for greater $S$ in terms of the Figure 3 right hand side plot is to be expected since the algorithm does not explicitly search for the most likely sequence). The required number of paths is significantly more than reported by Sutskever et al. (2014), who observed marginal gains for $S > 2$. Broadly speaking, this difference is to be expected: locally, the best musical harmonisation may be highly ambiguous, requiring many paths to be maintained in order for the future fixed (or observed) notes to have a reasonable chance of resolving well in the musical sense. The machine translation problem considered by Sutskever et al. (2014) is less ambiguous, and furthermore future symbols (words) are not conditioned on as in our setting.

|     | Hyper-link      | Description                      |
| --- | --------------- | -------------------------------- |
| (a) | melody          | original melody line in isolation |
| (b) | original        | complete original composition    |
| (c) | beam search     | melody line + beam search        |
| (d) | particle filter | melody line + particle filter    |

Table 1: Sample output based on the first movement of the *Quartet No. 7 in E flat major* by Mozart. In (a) the original melody line is rendered by itself with a piano timbre. In (b)—(d), three additional parts are rendered with a distinct classical guitar like timbre.

### 4.3. Human Listening Test

We also investigated the effectiveness of our algorithms as judged by human evaluators, via an on-line survey using the *Amazon Mechanical Turk*[3].

#### 4.3.1. METHODOLOGY

As the basis for the survey, we generated harmonisations as in the previous subsection 4.2, with the number of candidate paths $S$ set to 4096. We excluded two of the sixteen pieces previously considered, due to the extended use of pure pedal point (repeating of a single note) in three out of four voices *of the original compositions*, leading to uninteresting constraints in the majority of cases. For each of the remaining fourteen pieces and for $m \in \{0, 1, 2, 3\}$, we harmonised the pieces with $m$ out of four voices fixed. As before, we did this all $C_m^4$ possible ways, for both the particle filter and beam search methods. Each such output was compared with the original piece from which it was derived. This comparison was performed by ten unique human subjects per piece (not the same ten for each piece; there were 67 unique respondents). This gave a grand total of $14 \times \sum_{n=0}^{3} C_n^4 \times 2 \times 10 = 4200$ comparisons, or — figuring conservatively at one minute per test — approximately 72 hours of listening. Each experiment presented the original and derived piece in random order, and asked the participant to decide *which sounds better overall?*

To improve data quality we included two participant filters. First, we presented pieces with random notes drawn from uniform distribution over a four octave range, and excluded subjects who failed to do better than random at selecting the original piece over the random piece. Second, we tried to filter insincere subjects who responded in a very short time.

#### 4.3.2. FINDINGS

A summary of the results is presented in Figure 4, which presents frequentist confidence intervals on the unknown binomial probability (Brown et al., 2002) of choosing our pieces as preferable to the original. Note that a probability (vertical axis) of greater than one half would imply that our results are on average preferable to the original piece.

---

3. http://www.mturk.com

| Retained Part | Particle Filter | Beam Search |
| --- | --- | --- |
| 4 (melody) | hyperlink ($\sim$ d) | hyperlink (c) |
| 3 | hyperlink | hyperlink |
| 2 | hyperlink | hyperlink |
| 1 (bass) | hyperlink | hyperlink |

Table 2: Sample output based on the first movement of the *Quartet No. 7 in E flat major* by Mozart. Similarly to Table 1 (please see the corresponding caption), we retained a single part as composed by Mozart, and replaced the pitches in the remaining 3 voices. *Retained Part*: the part which we left unchanged from the original, from 4 (highest pitch) to 1 (lowest); *Particle Filter:* the sampling result; *Beam Search:* the beam search result. (c) is identical to that of Table 1, while ($\sim$ d) is similar to (d) from Table 1, but used a different random seed.

The most clear result is that with no fixed parts, the beam search algorithm tends to produce inferior results. This is in line with the experience of our informal experimentation with the beam search approach which, in the absence of fixed notes to condition on, tended to produce overly regular patterns featuring long sequences of repeated notes, *etc*. This is not an erroneous result, but rather a natural consequence of maximising the sequence probability, rather than sampling from the distribution.[4] It is worth noting that this overly regular behaviour of the beam search may be observed not only in that case of unconditional sampling, but also in pieces where the parts being conditioned on feature sufficiently protracted periods of inactivity, during which the beam search is relatively unconstrained (see the illustrative examples in subsection 4.4).

The particle filter does only slightly worse than parity with no fixed parts, but this effect is not significant based on these tests. Our own informal experimentation indicated that the fully unconstrained (no fixed parts) particle filter tends to occasionally finish pieces with poorly resolved harmonic movements — an observation partially corroborated by Figure 4. Interestingly, both algorithms perform well given even a single fixed part to condition on (with the particle filter performing slightly but consistently better overall). This is a broadly satisfactory result, which indicates the possibility of constructing convincing musical results by specifying a single voice (out of four), along with the rhythmic structure of the piece.

### 4.3.3. Discussion

One goal of our work is to create tools which facilitate the advancement of the musical art form. We believe that this may be be possible, through creative manipulation of the constraint set by skilled humans. Ideally, we would leverage the super human ability of the computer to sift through large numbers of possible solutions. In this way, constraint sets

---

4. Informal thought experiment: consider the stochastic relationship $y_i = x_i + \mathcal{N}(0, 1)$. We imagine the analogy that the $x_i$ represent the timing and the $y_i$ the pitch of a set of notes making up a piece. The most likely $y_i$ given an $x_i$ is exactly $y_i = x_i$, and yet for a set $\{x_i\}_{i=1}^n$ such a perfect straight line is in some sense highly atypical of a sample from the conditional $\{y_i\}_{i=1}^n | \{x_i\}_{i=1}^n$.

**Algorithm 1:** Sequential Monte Carlo for constrained non-Markov sequences, with systematic re-sampling (Carpenter et al., 1999).

**function** ConstrainedSMC($S, \{(f_i, h_i)\}_{i=1}^n$):
    **for** $s \leftarrow 1, 2, \ldots, S$ **do**
        $\boldsymbol{x}^{(s)} \leftarrow ()$         *initialise with empty sequence*
    **end**
    **for** $i \leftarrow 1, 2, \ldots, n$ **do**
        **for** $s \leftarrow 1, 2, \ldots, S$ **do**
            sample $x_i^{(s)} \sim q^{(\mathcal{A})}(\cdot \mid x_{1:i-1}^{(s)})$     *proposal (7)*
            $\boldsymbol{x}^{(s)} \leftarrow (\boldsymbol{x}^{(s)}, x_i^{(s)})$     *append*
            $w_s \leftarrow \sum_{\tilde{x}_i \in \mathcal{X}} f_i(\tilde{x}_i | x_{1:i-1}^{(s)}) h_i(\tilde{x}_{1:i}^{(s)})$     *(8)*
        **end**
        $k_{1:S} = \text{SystematicResample}(w_{1:S})$
        $(\boldsymbol{x}^{(s)})_{s=1}^S \leftarrow (\boldsymbol{x}^{(k_s)})_{s=1}^S$     *duplicate/delete particles*
    **end**
    **return** $\boldsymbol{x}^{(1:S)}$     *approximate samples $\boldsymbol{x}^{(1:S)}$ from $p^{(\mathcal{A})}$ of (6)*

**function** SystematicResample($w_{1:S}$):
    $\omega_{1:S} \leftarrow w_{1:S} / \sum_{s=1}^S w_s$
    sample $u \sim \text{Uniform}([0, 1])$
    $\bar{u} \leftarrow u/S; \quad j \leftarrow 1; \quad S_\omega \leftarrow \omega_1$
    **for** $l \leftarrow 1, 2, \ldots, S$ **do**
        **while** $S_\omega < \bar{u}$ **do**
            $j \leftarrow j + 1$
            $S_\omega \leftarrow S_\omega + \omega_j$
        **end**
        $k_l \leftarrow j$
        $\bar{u} \leftarrow \bar{u} + 1/S$
    **end**
    **return** $k_{1:S}$

which are too complex and tightly coupled for human investigation may yield new musical forms. Evaluating such a scheme is beyond the present scope, likely requiring an entirely different approach than the simple listening test undertaken here (see Loughran and O'Neill (2016) for a discussion of the challenges). These experiments are intended merely to verify that our system can produce feasible solutions given a simplistic constraint set.

### 4.4. Illustrative Example

Finally, in Table 1 we provide sample audio output from both the particle filter and beam search algorithms. To demonstrate the nature of the algorithms, we took a string quartet by Mozart, fixed the melody line, and re-harmonised the three remaining parts given their original rhythmic structure (note onset and offset times). The results are typical of the behaviour of the algorithms. In particular, the beam search produces more repetitive results than the particle filter.

To demonstrate robustness, Table 2 provides further re-harmonisations. The methodology is identical to that of the previous paragraph; in this case however we vary which voice was fixed to the original composition of Mozart, while re-harmonising the remaining three.

## 5. Conclusions

We presented algorithms for combining sophisticated probabilistic models of polyphonic music with human input. We represent human input as a finite state machine which accepts allowed compositions. Such a constraint is rather general, and yet amenable both to sequential Monte Carlo (for sampling from implied conditional distribution) and beam search optimisation (for maximising the same conditional distribution). We demonstrated the efficacy of the methods both quantitatively and through a listening experiment with human subjects. When the constraints are highly permissive, it seems that conditional sampling should be preferred to probability maximisation, as the latter tends to produce overly regular results in this case. Furthermore, the two approaches lead to a different style of musical result, as exemplified by the included audio example.